\documentclass[11pt]{article}

\usepackage[letterpaper,margin=1in]{geometry}
\usepackage[T1]{fontenc}
\usepackage[utf8]{inputenc}
\usepackage{lmodern}
\usepackage{microtype}
\usepackage{amsmath,amssymb,amsthm}
\usepackage{bm}
\usepackage{booktabs}
\usepackage{multirow}
\usepackage{makecell}
\usepackage{graphicx}
\usepackage[table]{xcolor}
\usepackage{xspace}
\usepackage{caption}
\usepackage{subcaption}
\usepackage{enumitem}
\usepackage{tikz}
\usetikzlibrary{arrows.meta,positioning,fit,backgrounds,calc,shapes.geometric}
\usepackage[ruled,vlined,linesnumbered]{algorithm2e}
\usepackage[numbers,sort&compress]{natbib}
\usepackage[colorlinks,allcolors=blue!55!black,breaklinks]{hyperref}
\usepackage[capitalise,nameinlink]{cleveref}

\setlist{itemsep=2pt,topsep=3pt}
\newtheorem{proposition}{Proposition}

\newcommand{\method}{CoMET\xspace}
\newcommand{\apseven}{AP@0.7\xspace}
\newcommand{\apfive}{AP@0.5\xspace}

\newcommand{\ie}{i.e.,\xspace}



\title{\method: Tracking the Shared Pose Error of Roadside Messages\\for Asynchronous Vehicle--Infrastructure Cooperative 3D Detection}
\author{Shuwei Sun \qquad Jian Huang \qquad Hui Cao\\[3pt]
  \small Xi'an Jiaotong University}
\date{}
\hypersetup{
  pdftitle={CoMET: Tracking the Shared Pose Error of Roadside Messages for Asynchronous Vehicle-Infrastructure Cooperative 3D Detection},
  pdfauthor={Shuwei Sun, Jian Huang, Hui Cao}
}

\begin{document}
\maketitle

\begin{abstract}
A roadside unit can see the car that a truck hides from the ego vehicle, but in deployment its message arrives late and can be placed only with poses that are available online.
We study vehicle--infrastructure cooperative 3D detection on V2X-Seq under a strict arrival protocol that admits only messages captured before the decision time and uses no ground truth, future frames or per-pair calibration offsets at inference.
In this setting most of the value of cooperation is lost to geometry rather than to latency: late fusion with online poses improves \apseven over the ego detector by only 1.0 point and intermediate fusion by at most 0.8, whereas the same late fusion placed with the dataset's annotation-assisted offsets would improve it by 7.4.
The loss is a slowly drifting translation shared by all objects of a message. The objects seen by both agents reveal it, but 38\% of messages contain no such object.

We propose \method (Common-Mode Error Tracking), which treats this translation as a latent state of the message stream: every ego detection is an anchor candidate; one observation per message is formed by mutual-nearest-neighbour Hungarian matching; a causal scalar Kalman filter tracks the offset per sender; and the estimate corrects every remote box, including those of objects the ego vehicle cannot see.
The estimator has no learned parameters, and messages carry only boxes, scores and capture times; a 1.3K-parameter readout head, trained once and frozen, re-ranks the merged candidates.

\method reaches 66.63 \apseven, 3.45 points above the same late fusion with online poses and 3.67 above the best intermediate fusion.
It improves all five detectors we test, from three architectures, by 2.2--3.5 points over online poses and by 0.3--0.6 over estimating the offset from the current message alone.
Nine variants of three per-frame registration modules, with and without temporal smoothing, remain 2.3--3.1 points behind.
We also find that the benchmark's reference offset is an exact-zero placeholder on 28\% of validation messages and comes from a bounded discrete search that saturates on 45\% of the remaining ones; because the roadside-only ground truth is built from this offset, the official evaluation inherits both artefacts.
We therefore report results on label-defined populations and under two auxiliary references; \method improves on every population under both.
Finally, retraining an intermediate-fusion network with \method poses and cleaned labels improves over retraining with online poses on all three architectures we test, significantly on two, and gives the highest \apfive of all methods.
\end{abstract}

\section{Introduction}
\label{sec:intro}

\begin{figure}[!t]
  \centering
  \includegraphics[width=\linewidth]{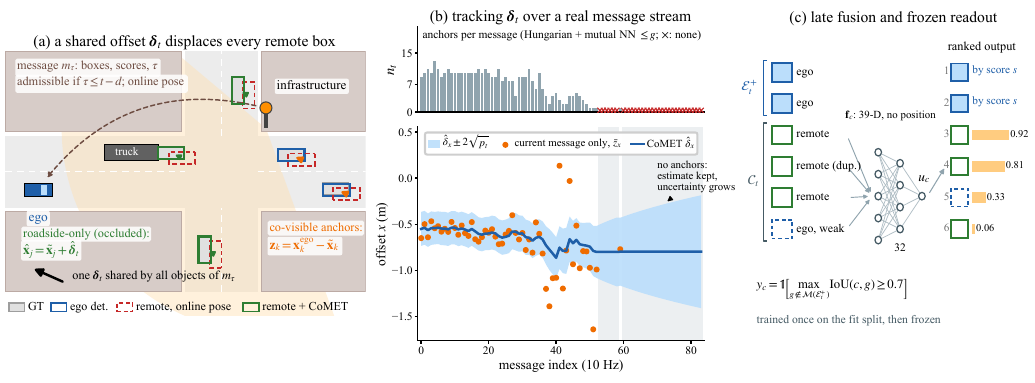}
  \caption{\textbf{\method at a glance.}
  (a)~A roadside message is admissible only if it was captured at least $d$\,ms before the decision time $t$. Placed with online poses, all remote boxes (red, dashed) are displaced from the ground truth (gray) by the same offset $\boldsymbol\delta_t$ (exaggerated here). Objects seen by both agents yield anchor residuals $\mathbf z_k$ between ego boxes (blue) and remote boxes that measure $\boldsymbol\delta_t$; the tracked estimate moves every remote box, including the roadside-only objects hidden behind the truck (green).
  (b)~A real validation stream (sequence 0071, $d=0$). Anchors per message come from Hungarian matching of all ego detections with a mutual-nearest-neighbour gate. The estimate from the current message alone (orange) is noisy and absent when there are no anchors (gray bands); the \method estimate (blue) is stable, is kept through anchor-free messages and its uncertainty grows until anchors return.
  (c)~Confident ego detections form a protected prefix $\mathcal E^+_t$; corrected remote boxes and weak ego boxes $\mathcal C_t$ are ranked by a 1.3K-parameter readout head on 39 receiver-side features without absolute position, trained once and frozen. Only boxes, scores and capture times are transmitted.}
  \label{fig:overview}
\end{figure}

An ego vehicle approaching an intersection cannot see the car hidden behind a truck; a LiDAR mounted on a pole above the intersection can.
Telling the vehicle where that car is, early enough to plan around it, is the promise of vehicle--infrastructure (V2I) cooperative perception~\cite{yu2022dairv2x,yu2023v2xseq}.
Most methods fuse the two views at the feature level~\cite{chen2019fcooper,wang2020v2vnet,li2021disconet,xu2022v2xvit,hu2022where2comm,xu2022cobevt} or at the box level~\cite{yu2022dairv2x}, and most reported gains are measured with views that are synchronised and registered with carefully calibrated poses.

A deployed receiver has neither.
The roadside message was captured before the ego frame and arrives after a communication delay, and it can only be placed with the poses that are available online: the vehicle's localisation estimate and the roadside calibration.
We evaluate V2X-Seq~\cite{yu2023v2xseq} under a \emph{strict arrival protocol} that admits only messages captured at least $d$\,ms before the decision time ($d\in\{0,100,200,400\}$) and that never uses ground truth, future frames or the dataset's per-pair calibration offsets at inference.
Under this protocol, cooperation as currently practised adds little (\cref{fig:latency}): late fusion with online poses raises \apseven over the ego detector by 1.0 point, and three intermediate-fusion models change it by $-1.0$ to $+0.8$.
Yet the value is there: the same late fusion placed with the dataset's annotation-assisted offsets would raise \apseven by 7.4 points.
The decisive limitation is therefore geometric, not temporal.

\paragraph{The error is shared, persistent and sparsely observed.}
V2X-Seq provides one planar translation per frame pair, and applying it recovers most of the lost accuracy; a single translation shared by all objects of a message---a \emph{common-mode} error---therefore explains most of the misplacement.
This error is observable without any label: an object seen by both agents (an \emph{anchor}) yields a residual between the ego box and the projected roadside box that measures the shared translation, and the same correction then applies to the roadside-only objects that motivate cooperation in the first place.
A single message, however, carries little evidence.
On the validation sequences a message holds 2.96 anchors on average, and 38\% of messages hold none; estimates from the current message alone are therefore often impossible and, when possible, noisy (a radial error of 0.48\,m against a smoothed reference at 0\,ms and 0.61\,m at 400\,ms).
At the same time the error drifts slowly within a sequence (\cref{fig:trajectory}).
The misalignment is a property of the \emph{message stream}, not of any individual frame.

\paragraph{Tracking instead of registering.}
Latency-aware methods~\cite{lei2022syncnet,wei2023cobevflow,yu2023ffnet,yang2023how2comm,song2025trafalign} compensate motion between capture and use and assume that the relative pose is right.
Pose-robust methods~\cite{lu2023coalign,lei2024freealign,huang2024roco,song2023cbm} re-register each frame from its own correspondences and are mostly evaluated under independent synthetic noise.
Neither treats the partner's error as persistent: the first leaves it uncorrected, and the second re-estimates it from scratch in every frame, so no estimate is available when correspondences are missing.
We propose \method (Common-Mode Error Tracking), which treats the partner's current systematic error as a state to be estimated together with its uncertainty.
\method uses every ego detection as an anchor candidate, forms one robust observation per message with mutual-nearest-neighbour Hungarian matching, tracks the shared offset with a causal Kalman filter per sender, carries the estimate through messages without anchors, and transfers it to every remote box (\cref{fig:overview}).
A small frozen readout head then re-ranks the corrected boxes together with the low-score ego detections.
Because process and measurement noise are isotropic, the filter reduces exactly to a scalar recursion with a two-coordinate state and one variance per stream; it has no learned parameters and needs only the boxes that a late-fusion message already carries.
Its two limiting cases are the two classical answers to pose error, per-frame registration and fixed calibration, and on real data it outperforms both.

\paragraph{The benchmark's reference is itself incomplete.}
Evaluating such a method requires care, because the reference offsets of V2X-Seq are not what they appear to be.
We audit them against the raw annotations and find that (i) the offset is \emph{exactly zero} on 28.0\% of validation messages, nearly all of which contain no co-labelled object, so zero is a placeholder rather than a measurement; (ii) the roadside-only ground truth is the public projection plus this offset (residual $\approx4\times10^{-7}$\,m), so on placeholder frames the ground truth itself is uncorrected; and (iii) the non-zero offsets come from a bounded discrete search with 26 values per axis whose bound is reached on 44.6\% of non-zero validation messages, where the co-labelled objects are typically displaced beyond the bound (\cref{fig:audit}).
An estimator that is right on a placeholder frame is therefore penalised by the official evaluation, and the dataset offset is a label-consistent reference rather than an oracle.
We report the official evaluation as the primary result and complement it with label-defined populations and two auxiliary references; under both auxiliary references \method improves on every population.

\paragraph{Results.}
With the boxes of the released TraF-Align detector, \method reaches 66.63 \apseven averaged over the four latency budgets, 3.45 points above late fusion with online poses and 3.67 above the best intermediate fusion, while keeping the advantage at every budget.
It improves the late fusion of all five detectors we test by 2.2--3.5 points over online poses and by 0.3--0.6 over estimating the offset from the current message alone. Nine variants of three per-frame registration modules, with and without temporal smoothing, remain 2.3--3.1 points behind, and as the pose input of CoAlign, \method beats CoAlign's own pose graph by 1.1 points.
When intermediate fusion is retrained, using \method poses and cleaned labels is better than retraining with online poses on all three architectures we test (significantly on two), and gives the highest \apfive of all methods on the strongest one.

\paragraph{Contributions.}
\begin{itemize}
  \item We show that under strict arrival and online poses the main obstacle to real V2I cooperative detection is a slowly drifting common-mode translation, which co-visible objects reveal but which no single message estimates reliably.
  \item We propose \method, a training-free causal tracker of this translation with cross-object transfer, which interpolates between per-frame registration and fixed calibration and costs a single small assignment per message.
  \item We audit the reference offsets of V2X-Seq, identify a zero placeholder and a saturated bounded search, explain how they enter the official ground truth, and propose an evaluation that separates label artefacts from method effects.
  \item We provide an extensive evaluation with paired sequence-level confidence intervals, hyper-parameters selected on a separate calibration split only, every reported number re-derived by an independent implementation, and a retraining study across three architectures.
\end{itemize}

\section{Related Work}
\label{sec:related}

\paragraph{Cooperative 3D detection.}
Cooperative perception exchanges raw point clouds (early fusion), intermediate features, or detected boxes (late fusion).
Intermediate fusion dominates recent work: F-Cooper~\cite{chen2019fcooper} max-pools voxel features, V2VNet~\cite{wang2020v2vnet} aggregates them with a graph network, DiscoNet~\cite{li2021disconet} distils an early-fusion teacher, V2X-ViT~\cite{xu2022v2xvit} and CoBEVT~\cite{xu2022cobevt} attend across agents, and When2com~\cite{liu2020when2com} and Where2comm~\cite{hu2022where2comm} reduce bandwidth through selective communication.
Benchmarks include OPV2V~\cite{xu2022opv2v} in simulation, DAIR-V2X~\cite{yu2022dairv2x} and V2X-Seq~\cite{yu2023v2xseq} for real V2I, and V2V4Real~\cite{xu2023v2v4real} for real V2V.
Late fusion suits scarce bandwidth and heterogeneous detectors but is generally considered weaker~\cite{yu2022dairv2x}.
We find that, once the shared geometric error is tracked, box-level fusion outperforms the intermediate-fusion models we test at the strict IoU threshold, and that the same pose estimate also helps when intermediate fusion is retrained.

\paragraph{Asynchrony and latency.}
A partner's message is captured at another time and arrives with delay.
SyncNet~\cite{lei2022syncnet} compensates latency with a feature-level temporal model, CoBEVFlow~\cite{wei2023cobevflow} transports BEV features along estimated object flow, FFNet~\cite{yu2023ffnet} transmits feature flow so that the receiver can predict the roadside feature at the ego time, How2comm~\cite{yang2023how2comm} couples flow-guided compensation with communication selection, TraF-Align~\cite{song2025trafalign} aligns features along predicted trajectories, and CoDynTrust~\cite{xu2025codyntrust} modulates features with a dynamic trust estimate.
These methods model the \emph{motion} between capture and use and assume a correct relative pose.
On real V2I data with online poses we find the larger error to be geometric and persistent; \method addresses it at the box level and is complementary to motion compensation.

\paragraph{Pose errors in cooperative perception.}
V2X-ViT~\cite{xu2022v2xvit} learns robustness to injected pose noise; CoAlign~\cite{lu2023coalign} optimises an agent--object pose graph built from predicted boxes and their uncertainties in every frame; FreeAlign~\cite{lei2024freealign} matches salient-object graphs to recover the relative pose without external localisation; RoCo~\cite{huang2024roco} alternates object matching and pose adjustment; CBM~\cite{song2023cbm} matches objects through their local context; LET-VIC~\cite{yang2024letvic} compensates calibration errors inside a BEV cross-attention module for end-to-end V2I tracking.
Apart from the learned LET-VIC module, these methods estimate the relative pose \emph{from the current frame}, and they are mostly evaluated with independent Gaussian perturbations of synchronous poses.
\method instead exploits the temporal persistence of the error, accumulates evidence over the message stream with a causal filter, and transfers the estimate to objects without any correspondence in the current frame.
We compare with FreeAlign, RoCo and CBM, with and without temporal smoothing of their outputs, under an identical readout, and we feed \method poses to CoAlign.

\paragraph{Temporal filtering and online calibration.}
Kalman filtering~\cite{kalman1960} underlies tracking-by-detection~\cite{weng2020ab3dmot,yin2021centerpoint}, where each object carries its own state. Iterative registration~\cite{besl1992icp} recovers a rigid transform from dense correspondences between a single pair of scans.
\method filters a single \emph{shared} state per partner, the offset common to all of its objects, which is closer in spirit to online extrinsic calibration than to object tracking, and it estimates this state from sparse, noisy correspondences accumulated over time.
It needs neither raw point clouds nor calibration targets: everyday traffic seen by both agents serves as the calibration target.

\paragraph{Reference quality of cooperative benchmarks.}
Cooperative datasets provide calibration and pose refinements that are produced offline, and models are commonly trained and evaluated with them~\cite{yu2022dairv2x,yu2023v2xseq,xu2023v2v4real}.
We are not aware of an audit of how such refinements are generated and how they enter the cooperative ground truth.
Our analysis of V2X-Seq shows that both questions matter for evaluating any method that corrects geometry online.

\section{Problem: Where the Value of Cooperation Is Lost}
\label{sec:problem}

\subsection{Setting}
\label{sec:setting}
An ego vehicle $e$ and a roadside unit $i$ each run a LiDAR detector.
At decision time $t$ the ego detector outputs boxes $\mathcal B_e(t)=\{(\mathbf b_k,s_k)\}$, and the roadside unit broadcasts messages $m_\tau=(\mathcal B_i(\tau),\tau)$ with its detections at capture time $\tau$.
Under the \emph{strict arrival protocol} the receiver may use only messages with $\tau\le t-d$ for a latency budget $d$; frames without an admissible message are still evaluated.
Remote boxes are mapped into the ego frame with the poses available online---the vehicle localisation $\mathbf T_{w\leftarrow e}(t)$ and the roadside calibration $\mathbf T_{w\leftarrow i}$---so that $\mathbf T_{e\leftarrow i}(t)=\mathbf T_{w\leftarrow e}(t)^{-1}\mathbf T_{w\leftarrow i}$.
No ground truth, future frame or per-pair calibration offset is used at inference.
The task is to detect vehicles in the ego frame at time $t$.

\subsection{Geometry, not latency, dominates}
\label{sec:geometry}
\begin{figure}[t]
  \centering
  \includegraphics[width=0.78\linewidth]{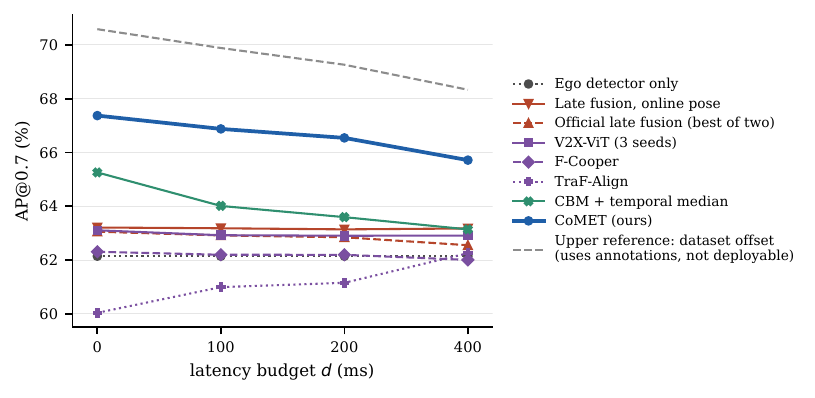}
  \caption{\textbf{\apseven against the latency budget on V2X-Seq validation.} Intermediate fusion and late fusion with online poses stay close to the ego detector. \method recovers over half of the gap to the upper reference that places the boxes with the dataset's annotation-assisted offset (dashed gray, not deployable). Values are means over three readout heads or three detector seeds; numbers in \cref{tab:main}.}
  \label{fig:latency}
\end{figure}

\Cref{fig:latency} shows \apseven on the 21 validation sequences against the latency budget (exact numbers in \cref{tab:main}).
Three observations frame the rest of the paper.
\emph{(i) Cooperation barely helps as practised.}
Late fusion of the released TraF-Align~\cite{song2025trafalign} boxes with online poses improves \apseven over the ego detector by 1.0 point, to 63.18 on average, and the official late-fusion baselines and three intermediate-fusion models stay between 61.1 and 63.0.
\emph{(ii) The value is recoverable.}
Placing the same remote boxes with the dataset's annotation-assisted offset, one planar translation per frame, gives 69.52 \apseven: one translation per message adds 6.3 points over online poses and turns a 1.0-point gain of cooperation into a 7.4-point gain.
\emph{(iii) Latency costs little in comparison.}
With good geometry, \apseven drops by 2.3 points from 0 to 400\,ms with the dataset offset (70.59 to 68.33) and by 1.7 points with \method (67.37 to 65.72), well below the geometric gap.

\subsection{A shared, persistent, sparsely observed error}
\label{sec:structure}
We model the centre of remote object $j$ after mapping to the ego frame as
\begin{equation}
  \tilde{\mathbf x}_j=\mathbf x_j-\boldsymbol\delta_t+\boldsymbol\epsilon_j ,
  \label{eq:model}
\end{equation}
where $\mathbf x_j$ is its true centre, $\boldsymbol\delta_t$ a planar correction shared by all objects of the message (the common mode) and $\boldsymbol\epsilon_j$ an object-specific error from detection noise and residual motion.
Such a shared term arises when the dominant error lies in the transform that maps the roadside frame into the ego frame---a residual calibration or localisation bias---which displaces every object of a message alike, whereas detection noise and uncompensated motion act on each object separately.
Three properties make this model useful.

\emph{Observable.}
For an object seen by both agents, the residual $\mathbf z=\mathbf x^{\text{ego}}-\tilde{\mathbf x}$ between the ego box and the projected remote box equals $\boldsymbol\delta_t$ plus noise, with no label involved.
The same $\boldsymbol\delta_t$ then corrects the remote objects that the ego vehicle cannot see.

\emph{Sparse per message.}
Using all ego detections and mutual-nearest-neighbour matching within 2.5\,m (\cref{sec:method}), a validation message contains 2.96 anchors on average at $d=0$ (2.73 at 400\,ms), and 38.2\% of messages contain none; messages that do contain anchors hold 4.79 of them.
Estimating $\boldsymbol\delta_t$ from the current message alone is therefore often impossible and, when possible, noisy: against a smoothed reference built from the dataset offsets (only used for analysis), the per-message estimate has a radial RMSE of 0.48\,m at $d=0$ and 0.61\,m at $d=400$\,ms on messages with anchors, and misses by more than 1\,m on 4\% and 9\% of them; moving anchors increasingly mix the pose error with uncompensated motion as the message ages.
The tracked estimate of \method has 0.41\,m RMSE at both budgets and misses by more than 1\,m on 2\%.

\begin{figure}[t]
  \centering
  \includegraphics[width=\linewidth]{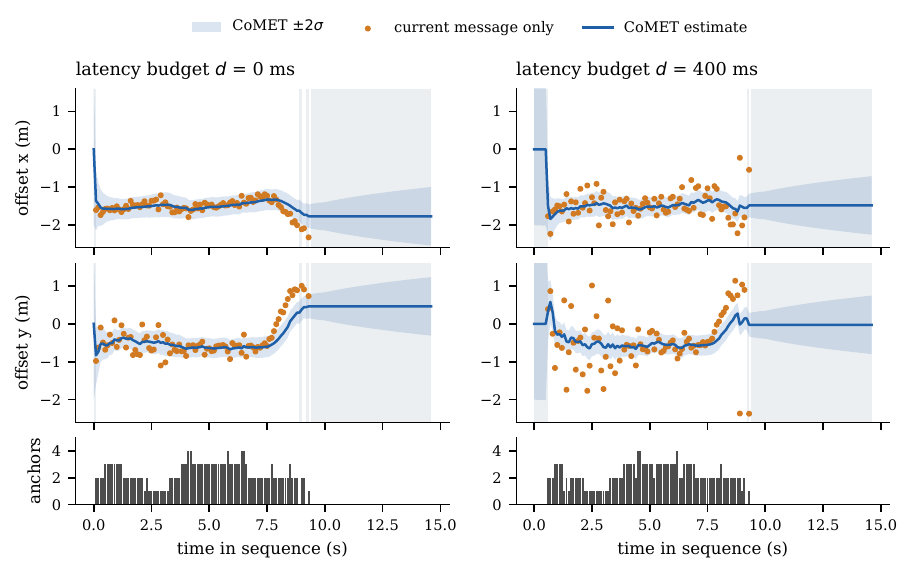}
  \caption{\textbf{One validation stream (sequence 0007) at two latency budgets.} Orange: offset estimated from the anchors of the current message only (shown only when the message has anchors). Blue: the \method estimate with its $\pm2\sigma$ band; gray bands mark messages without anchors. Bottom: number of anchors per message. The per-message estimate is noisy and disappears when anchors are absent (after about 9\,s); the tracked estimate is stable, follows the change in $y$ around 8\,s, and keeps the last estimate with growing uncertainty when anchors vanish. At 400\,ms the per-message estimate is noticeably noisier.}
  \label{fig:trajectory}
\end{figure}

\emph{Persistent.}
Within a sequence the offset changes slowly compared with the 10\,Hz message rate (\cref{fig:trajectory}).
The common-mode error should therefore be \emph{tracked} over the message stream rather than re-estimated in every frame.

\subsection{The benchmark's reference offset}
\label{sec:audit}
\begin{figure}[t]
  \centering
  \includegraphics[width=\linewidth]{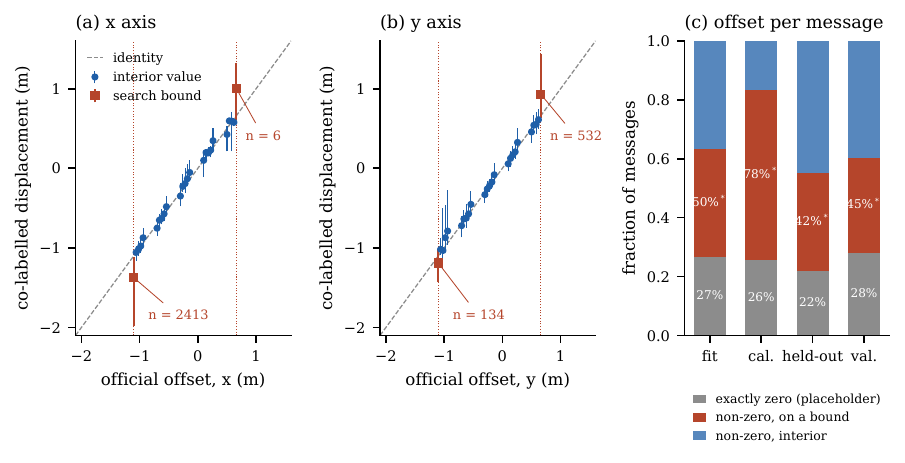}
  \caption{\textbf{The dataset offset is a bounded discrete search with a zero placeholder.} (a,b) For each value the offset takes on an axis, the median and 10--90\% range of the displacement of co-labelled objects in the same frame (frames with at least three co-labelled objects, all splits). Interior values follow the diagonal; on the bounds (red) the objects are displaced further than the offset. (c) Per split, the share of messages with an exactly zero offset (gray, percentage of all messages), a non-zero offset on a bound (red; the percentage marked $^*$ is relative to non-zero messages), and an interior non-zero offset (blue).}
  \label{fig:audit}
\end{figure}

V2X-Seq provides a per-pair \texttt{system\_error\_offset}, a planar translation estimated offline for each vehicle--roadside frame pair.
Prior work trains and evaluates with these offsets as the calibrated relative pose.
Because they also define what ``correct geometry'' means in \cref{sec:geometry}, we audit them directly against the raw annotations of all 67 training and validation sequences (\cref{tab:census}, \cref{fig:audit}; details in \cref{app:audit}).

\begin{table}[t]
  \centering
  \caption{\textbf{Census of the dataset offset per split.} Messages with an offset of exactly $(0,0)$; among them, those that still contain a co-labelled object; non-zero messages without a co-labelled object; and non-zero messages with at least one axis on the search bound ($-1.1$ or $+0.66$\,m), as a fraction of non-zero messages.}
  \label{tab:census}
  \small
  \begin{tabular}{lrrrrr}
    \toprule
    Split & Messages & Exactly zero & \makecell{Zero with\\co-labelled obj.} & \makecell{Non-zero without\\co-labelled obj.} & \makecell{Non-zero\\on a bound} \\
    \midrule
    fit (24 seq.) & 3762 & 1001 (26.6\%) & 0 & 16 & 1379 (49.9\%) \\
calibration (8) & 1247 & 319 (25.6\%) & 35 & 1 & 721 (77.7\%) \\
held-out (14) & 2436 & 530 (21.8\%) & 0 & 10 & 810 (42.5\%) \\
validation (21) & 3316 & 928 (28.0\%) & 0 & 11 & 1066 (44.6\%) \\
 
    \bottomrule
  \end{tabular}
\end{table}

\paragraph{A placeholder.}
The offset is exactly zero on 21.8--28.0\% of the messages of each split.
Outside the calibration split, no zero message contains a co-labelled object (an object annotated by both agents), and only 10--16 of the messages without a co-labelled object have a non-zero offset.
Zero is therefore a placeholder for ``could not be estimated'', not a measurement of zero error.
The calibration split contains 35 zero messages that do have a co-labelled object; there the roadside and vehicle labels of that object differ by 1.22\,m (median), so the placeholder is not a statement that the error vanished.

\paragraph{The offset is inside the ground truth.}
For roadside-only objects, the cooperative ground truth equals the public projection of the roadside label plus the offset, with a residual of about $4\times10^{-7}$\,m in both the zero and the non-zero case.
On placeholder messages the ground truth of roadside-only objects is therefore the \emph{uncorrected} projection.
A method that estimates the true shift on such a message moves its boxes away from this ground truth and is penalised; a method that does nothing is rewarded.

\paragraph{A bounded discrete search.}
The non-zero offsets take only 26 distinct values per axis including zero, in five groups with a step of 0.04\,m, bounded by $-1.1$ and $+0.66$\,m (\cref{fig:audit}a,b).
At least one axis sits on a bound for 44.6\% of the non-zero validation messages (77.7\% in the calibration split).
On the bound the co-labelled objects of the same frame are typically displaced further: on the $x=-1.1$ bound (2,413 frames with at least three co-labelled objects) their median displacement is $-1.37$\,m and 91\% of the frames lie beyond the bound; on the $y=+0.66$ bound the median is $+0.93$\,m.
Across frames with at least three co-labelled objects, displacements larger than 1.5\,m occur in 19.6\% of frames but official offsets in only 1.1\%.
We infer the search structure from the values alone; the generating procedure is not documented.

\paragraph{Consequences for evaluation.}
The dataset offset is a label-consistent reference, not an oracle: it is zero where it could not be estimated and clipped where the error is large, and the ground truth inherits both properties.
We keep the official ground truth for all primary results and add two complementary readings (\cref{sec:protocol}): results split into label-defined populations (all, exactly zero, with co-labelled objects), and two auxiliary references that re-register roadside-only ground truth on placeholder frames (by interpolating the non-zero offsets of the same sequence) or on all frames with at least three co-labelled objects (by their median displacement).
Neither auxiliary reference is physical truth; they show which conclusions depend on the label artefacts.

\section{\method: Common-Mode Error Tracking}
\label{sec:method}

\method runs on the receiver and treats both detectors as black boxes (\cref{fig:overview}, \cref{alg:comet}).
It builds remote candidates at the decision time (\cref{sec:prop}), forms one anchor observation per message (\cref{sec:anchors}), tracks the common-mode offset with a causal filter (\cref{sec:kf}), transfers it to every remote box and ranks the merged candidates (\cref{sec:readout}).
\Cref{sec:tier2} describes an optional second tier that retrains intermediate fusion with \method poses.

\subsection{Remote candidates at the decision time}
\label{sec:prop}
A message $m_\tau$ carries boxes $\mathbf b_j=(\mathbf x_j,l_j,w_j,h_j,\theta_j,s_j)$ in the roadside frame.
After per-message non-maximum suppression, the receiver associates $m_\tau$ with the previous admissible message $m_{\tau'}$ in world coordinates and propagates every associated box to the decision time,
\begin{equation}
  \tilde{\mathbf x}_j=\mathbf T_{e\leftarrow i}(t)\big(\mathbf x_j^{\tau}+\mathbf v_j\,(t-\tau)\big),\qquad
  \mathbf v_j=\frac{\mathbf x_j^{\tau}-\mathbf x_{a(j)}^{\tau'}}{\tau-\tau'},
  \label{eq:prop}
\end{equation}
where $a(j)$ is the associated box of the previous message and unassociated boxes keep $\mathbf v_j=\mathbf 0$.
In \cref{eq:prop} we use the heading-constrained velocity $(\mathbf h_j^{\top}\mathbf v_j)\,\mathbf h_j$, with $\mathbf h_j=(\cos\theta_j,\sin\theta_j)^{\top}$, which suppresses lateral velocity noise.
Only past messages are used.

\subsection{Anchors and the per-message observation}
\label{sec:anchors}
Every ego detection is an anchor candidate; we apply no score threshold (\cref{sec:ablation} shows that thresholds only remove evidence).
Let $\mathbf X^{\text{ego}}$ be the ego box centres and $\tilde{\mathbf X}$ the propagated remote centres of the current message.
We solve the assignment $\pi^\star=\arg\min_\pi\sum_k\|\mathbf x^{\text{ego}}_k-\tilde{\mathbf x}_{\pi(k)}\|$ with the Hungarian algorithm~\cite{kuhn1955hungarian} and keep a pair only if it is within a gate $g$ and the two boxes are each other's nearest neighbour:
\begin{equation}
  \mathcal A_t=\big\{(k,j):\ j=\pi^\star(k),\ j=\mathrm{NN}(k),\ k=\mathrm{NN}(j),\ \|\mathbf x^{\text{ego}}_k-\tilde{\mathbf x}_j\|\le g\big\}.
  \label{eq:anchors}
\end{equation}
By \cref{eq:model}, an anchor residual observes the common mode directly,
\begin{equation}
  \mathbf z_{kj}=\mathbf x^{\text{ego}}_k-\tilde{\mathbf x}_j=\boldsymbol\delta_t+\boldsymbol\eta_{kj},\qquad\boldsymbol\eta_{kj}\sim\mathcal N(\mathbf 0,r^2\mathbf I),
  \label{eq:obs}
\end{equation}
where $\boldsymbol\eta_{kj}$ lumps the ego and remote object-specific errors.
For $n_t=|\mathcal A_t|$ anchors, the mean residual $\bar{\mathbf z}_t$ is a sufficient statistic, $\bar{\mathbf z}_t\sim\mathcal N(\boldsymbol\delta_t,\tfrac{r^2}{n_t}\mathbf I)$.
Matching is done on the propagated centres \emph{before} any correction, so a large current error cannot be hidden by a wrong prediction; using the filter prediction to gate the association (``prediction gating'') is worse in our experiments (\cref{tab:ablation}).
The mutual-nearest-neighbour test removes most ambiguous pairs in dense traffic.

\subsection{Tracking the common mode}
\label{sec:kf}
The offset persists but drifts, which we model as a random walk over the arrivals of admissible messages from one sender, $\boldsymbol\delta_t=\boldsymbol\delta_{t-1}+\mathbf w_t$, $\mathbf w_t\sim\mathcal N(\mathbf 0,q^2\mathbf I)$.
Together with \cref{eq:obs} this is a linear-Gaussian model, so the posterior given all past anchors is Gaussian.

\begin{proposition}[Scalar recursion]
\label{prop:scalar}
If the prior covariance, the process noise and the measurement noise are all isotropic, the posterior covariance stays isotropic, $\mathbf P_t=p_t\mathbf I$, and the Kalman filter reduces exactly to
\begin{align}
  p_t^-&=p_{t-1}+q^2, &
  k_t&=\begin{cases}\dfrac{p_t^-}{p_t^-+r^2/n_t}, & n_t\ge1,\\[1ex] 0, & n_t=0,\end{cases}\label{eq:gain}\\
  \hat{\boldsymbol\delta}_t&=\hat{\boldsymbol\delta}_{t-1}+k_t\big(\bar{\mathbf z}_t-\hat{\boldsymbol\delta}_{t-1}\big), &
  p_t&=(1-k_t)\,p_t^- .\label{eq:update}
\end{align}
\end{proposition}
The proof is a two-line induction (\cref{app:proof}).
Each stream starts from $\hat{\boldsymbol\delta}_0=\mathbf 0$ and $p_0=1$\,m$^2$.
A message with many anchors receives a large gain; a message without anchors leaves the estimate unchanged while $p_t$ grows; the first anchors after a gap are weighted strongly, so the estimate catches up quickly after an occlusion (\cref{fig:trajectory}).

\paragraph{Between registration and calibration.}
\Cref{eq:update} is an exponentially weighted average $\hat{\boldsymbol\delta}_t=(1-k_t)\hat{\boldsymbol\delta}_{t-1}+k_t\bar{\mathbf z}_t$ whose gain is set by $q/r$, the anchor count and the time since the last anchor.
Its two limiting cases are the classical answers to pose error: $q\to\infty$ gives $k_t\to1$ and $\hat{\boldsymbol\delta}_t=\bar{\mathbf z}_t$, \ie per-frame registration from the current correspondences; $q\to0$ gives $k_t\to0$ after convergence, \ie a fixed calibration averaged over the stream.
For a stream with $n$ anchors per message the gain converges to
\begin{equation}
  k_\infty=\frac{p^-_\infty}{p^-_\infty+r^2/n},\qquad p^-_\infty=\tfrac12\Big(q^2+\sqrt{q^4+4q^2r^2/n}\Big).
  \label{eq:steady}
\end{equation}
With $q=0.05$\,m, $r=0.6$\,m and the 4.79 anchors that an anchor-bearing validation message holds on average, $k_\infty\approx0.17$: the estimate pools roughly the last six anchor-bearing messages, about one second at 10\,Hz once the anchor-free messages are included.
That is long enough to average out object-level noise and short enough to follow the drift.

\paragraph{Hyper-parameters.}
We fix $r=0.6$\,m and $g=2.5$\,m for all detectors and select $q\in\{0.025,0.05,0.1,0.15\}$\,m per detector on the calibration split only (0.05\,m for the released TraF-Align boxes; \cref{app:impl}).
Streams are kept per sender; in our offline replay each latency budget is an independent replay, so the stream key is (sequence, budget).

\paragraph{Transfer.}
The tracked offset corrects \emph{every} remote box, $\hat{\mathbf x}_j=\tilde{\mathbf x}_j+\hat{\boldsymbol\delta}_t$, including the roadside-only objects that have no ego counterpart.
Because $\hat{\boldsymbol\delta}_t$ persists, messages without any anchor are corrected as well.
This is the essential difference from per-frame registration, which returns no correction whenever no correspondence is found and therefore gives up exactly on the frames in which the partner's evidence matters most.

\begin{algorithm}[t]
  \caption{\method on the receiver at decision time $t$}
  \label{alg:comet}
  \small
  \KwIn{ego boxes $\mathcal B_e(t)$; admissible messages $m_\tau,m_{\tau'}$; online pose $\mathbf T_{e\leftarrow i}(t)$; stream state $(\hat{\boldsymbol\delta},p)$}
  \KwOut{ranked detections $\mathcal D_t$}
  $\{\tilde{\mathbf b}_j\}\leftarrow$ propagate and map $m_\tau$ by \cref{eq:prop}\;
  \If(\tcp*[f]{once per message}){$m_\tau$ is new}{
    $p\leftarrow p+q^2$;\quad $\mathcal A\leftarrow$ anchors of all ego boxes by \cref{eq:anchors}\;
    \If{$n=|\mathcal A|\ge1$}{
      $\bar{\mathbf z}\leftarrow\frac1n\sum_{(k,j)\in\mathcal A}(\mathbf x^{\text{ego}}_k-\tilde{\mathbf x}_j)$;\quad $k\leftarrow p/(p+r^2/n)$\;
      $\hat{\boldsymbol\delta}\leftarrow\hat{\boldsymbol\delta}+k(\bar{\mathbf z}-\hat{\boldsymbol\delta})$;\quad $p\leftarrow(1-k)\,p$\;
    }
  }
  $\hat{\mathbf b}_j\leftarrow\tilde{\mathbf b}_j$ with centre $\tilde{\mathbf x}_j+\hat{\boldsymbol\delta}$ for all $j$\tcp*{transfer}
  $\mathcal E^+\leftarrow\{\mathbf b_k\in\mathcal B_e(t):s_k\ge\tau_p\}$;\quad $\mathcal C\leftarrow\{\hat{\mathbf b}_j\}\cup(\mathcal B_e(t)\setminus\mathcal E^+)$\;
  $u_c\leftarrow f_\phi(\mathbf f_c)$ for $c\in\mathcal C$\tcp*{frozen readout head}
  $\mathcal D_t\leftarrow\mathcal E^+$ ranked by $1+s_k$, followed by $\mathcal C$ ranked by $u_c$\;
\end{algorithm}

\subsection{Late fusion and readout}
\label{sec:readout}
Correct geometry is necessary but not sufficient: a corrected box only helps if it outranks the false positives it competes with, and the scores of two detectors are not comparable.
We therefore rank the confident ego detections $\mathcal E^+_t=\{k:s_k\ge\tau_p\}$ first and score every other candidate, \ie corrected remote boxes and low-score ego boxes, with a small frozen readout head.
The head is a two-layer perceptron with 32 hidden units (1.3K parameters) on 39 receiver-side features such as message age, box geometry and the overlap with the nearest ego detection; the 13 features that encode absolute position, including the sensor-to-sensor translation, are set to zero so that the head cannot learn location priors (\cref{app:impl}).
It is trained once with binary cross-entropy on the fit split to predict whether a candidate adds a correct detection (IoU $\ge0.7$ with a ground-truth box not yet matched by $\mathcal E^+_t$), and then kept fixed; it never sees the validation or held-out splits.
$\tau_p$ is selected per detector on the calibration split, and whether to keep the original heads or refit them on \method geometry is decided once, globally, on the calibration split (the original heads are kept).
The pose estimator itself has no learned parameters.

\paragraph{Cost.}
A message carries boxes, scores and the capture time.
Per message the receiver solves one assignment between a few dozen boxes, applies the closed-form update of \cref{eq:update} and evaluates the 1.3K-parameter head on each candidate; the state is two coordinates and one variance per sender.

\subsection{Second tier: retraining intermediate fusion}
\label{sec:tier2}
\method produces a causal relative pose, so it can also replace the online pose of an intermediate-fusion model.
We refer to the pipeline above, with its training-free pose estimate and frozen readout, as the first tier.
When an intermediate-fusion model is retrained anyway, we change two things and nothing else.
\emph{Pose input:} the fusion network receives the remote features warped with the \method estimate at training and at test time.
\emph{Training labels:} on placeholder messages of the fit split, the roadside-only ground truth is re-registered with the per-sequence linear interpolation of the non-zero dataset offsets, so that the network is not trained to reproduce the uncorrected projection; this uses labels and is applied to training only.
We also tested a variant that additionally moves roadside-only labels on saturated axes to the median co-labelled displacement; the calibration split preferred the simpler variant (\cref{app:tier2}).
Evaluation always uses the official ground truth.

\section{Experimental Protocol}
\label{sec:protocol}

\paragraph{Data and splits.}
We use the sequential V2I perception subset of V2X-Seq~\cite{yu2023v2xseq}.
Cooperative labels of the test split are not public, so all reported numbers are on the 21 official validation sequences (3,316 messages at each budget, 31,123 vehicle boxes), which no detector has seen during training and which are never used for any selection.
The 46 training sequences are divided into a \emph{fit} split (24) for readout heads and retraining, a \emph{calibration} split (8) that is the only split used for selecting any hyper-parameter, and a \emph{held-out} split (14) that we report as a secondary result.
The held-out split lies inside the detectors' training data; results on it are reported but never used to choose anything.

\paragraph{Protocol and metric.}
We use the strict arrival protocol of \cref{sec:setting} with $d\in\{0,100,200,400\}$\,ms.
We report bird's-eye-view (BEV) average precision at IoU 0.7 and 0.5 (\apseven, \apfive) for vehicles in the common region $x\in[0,96]$\,m, $|y|\le38.4$\,m of the ego frame, with predictions ranked globally over the split.
Unless stated otherwise a number is the mean over the four budgets and over the three frozen readout heads (seeds 2023/2024/2025) or three training seeds.

\paragraph{Statistics.}
Confidence intervals are 95\% whole-sequence paired bootstrap intervals with 2,000 replicates~\cite{efron1993bootstrap}: sequences are resampled with replacement, both methods are evaluated on the same draw, and draws are paired before averaging over heads, seeds and budgets.
We recomputed every reported number from the saved detections with a second, independently written implementation of box matching, AP and bootstrap; the two implementations agree to numerical precision.
Where a table lists per-head intervals we give the envelope of the three heads' intervals.

\paragraph{Populations and auxiliary references.}
Following \cref{sec:audit} we additionally split each result into \emph{all} messages (primary), messages with an exactly \emph{zero} offset, and messages with at least one \emph{co-labelled} object.
The populations are defined by annotations, used for reading only, and never as an input or for selection.
The two auxiliary references move only roadside-only ground-truth boxes, keep all predictions, rankings and the evaluation region unchanged, and are used as sensitivity analyses:
\emph{interpolated} (placeholder messages re-registered with the per-sequence interpolation of non-zero offsets) and \emph{label-median} (messages with at least three co-labelled objects re-registered with their median displacement; otherwise the official or interpolated offset).

\paragraph{Detectors and baselines.}
\emph{Detectors.} For late fusion we use the single-agent boxes of five detectors from three architectures: the released TraF-Align~\cite{song2025trafalign} weights, TraF-Align retrained under the strict protocol, two V2X-ViT~\cite{xu2022v2xvit} models trained with different seeds, and Where2comm~\cite{hu2022where2comm}.
Unless stated otherwise, results use the released TraF-Align boxes, the strongest single-agent detector (62.15 \apseven).
\emph{Intermediate fusion.} F-Cooper~\cite{chen2019fcooper} and V2X-ViT (three seeds) trained under the strict protocol, and TraF-Align with its released weights, all with online poses.
\emph{Late fusion.} The official DAIR-V2X late-fusion baseline~\cite{yu2022dairv2x} without and with its time compensation (we report the better of the two per column), and our late fusion with online poses and the same readout as \method.
\emph{Per-frame registration.} The official box-level modules of FreeAlign~\cite{lei2024freealign}, RoCo~\cite{huang2024roco} and CBM~\cite{song2023cbm} on the same candidate pool, each followed by the same readout as \method; we additionally smooth each module's per-frame estimate with a temporal median or an innovation-gated Kalman filter whose parameters are selected on the calibration split.
\emph{Pose graph.} CoAlign~\cite{lu2023coalign} trained on V2X-Seq under our protocol with three seeds.
In \cref{sec:geometry} we also show, as an upper reference that is not deployable, late fusion placed with the dataset offset, which uses annotations of the current frame.

\paragraph{Reproducibility.}
The estimator is re-implemented in a stand-alone package that reproduces the per-message offsets, variances and anchor counts behind all tables exactly (maximum absolute difference 0 over 205,202 messages, five detectors and four splits); the code, the protocol and all result tables are released.

\section{Results}
\label{sec:results}

\subsection{Main comparison}
\label{sec:main}

\begin{table}[t]
  \centering
  \caption{\textbf{V2X-Seq validation under strict arrival with online poses.} \apseven and \apfive (\%) per latency budget $d$ and their mean. $^\dagger$Box-level methods on the released TraF-Align boxes with the same readout as \method (three frozen heads). $^\ddagger$Temporal median of CBM's per-frame estimates; the best of six smoothed external variants on the calibration split.}
  \label{tab:main}
  \small
  \setlength{\tabcolsep}{3.6pt}
  \resizebox{\linewidth}{!}{%
  \begin{tabular}{l rrrr r rrrr r}
    \toprule
    & \multicolumn{5}{c}{\apseven} & \multicolumn{5}{c}{\apfive} \\
    \cmidrule(lr){2-6}\cmidrule(lr){7-11}
    Method & 0 & 100 & 200 & 400 & mean & 0 & 100 & 200 & 400 & mean \\
    \midrule
    Ego detector only & 62.15 & 62.15 & 62.15 & 62.15 & 62.15 & 70.93 & 70.93 & 70.93 & 70.93 & 70.93 \\
\midrule
F-Cooper~\cite{chen2019fcooper} & 62.31 & 62.20 & 62.20 & 62.01 & 62.18 & 77.33 & 77.10 & 76.85 & 76.48 & 76.94 \\
TraF-Align~\cite{song2025trafalign} & 60.04 & 61.00 & 61.16 & 62.26 & 61.11 & 75.11 & 75.37 & 75.18 & 75.41 & 75.27 \\
V2X-ViT~\cite{xu2022v2xvit} (3 seeds) & 63.11 & 62.93 & 62.91 & 62.91 & 62.96 & 78.31 & 77.98 & 77.64 & 77.32 & 77.81 \\
\midrule
Official late fusion~\cite{yu2022dairv2x} (best of two) & 63.06 & 62.92 & 62.85 & 62.55 & 62.84 & 74.23 & 74.03 & 73.84 & 73.07 & 73.79 \\
Late fusion, online pose$^\dagger$ & 63.21 & 63.18 & 63.14 & 63.18 & 63.18 & 74.45 & 74.33 & 74.25 & 74.06 & 74.27 \\
FreeAlign~\cite{lei2024freealign}$^\dagger$ & 64.09 & 63.39 & 63.29 & 63.18 & 63.49 & 75.16 & 73.75 & 73.73 & 73.38 & 74.01 \\
RoCo~\cite{huang2024roco}$^\dagger$ & 65.49 & 64.33 & 63.78 & 63.84 & 64.36 & 77.01 & 75.25 & 74.09 & 74.02 & 75.09 \\
CBM~\cite{song2023cbm}$^\dagger$ & 65.26 & 64.10 & 63.67 & 63.21 & 64.06 & 76.43 & 74.82 & 73.72 & 72.85 & 74.46 \\
CBM + temporal median$^{\dagger\ddagger}$ & 65.26 & 64.01 & 63.60 & 63.14 & 64.00 & 76.58 & 74.82 & 73.66 & 72.94 & 74.50 \\
\midrule
\textbf{CoMET (ours)}$^\dagger$ & \textbf{67.37} & \textbf{66.88} & \textbf{66.54} & \textbf{65.72} & \textbf{66.63} & \textbf{79.55} & \textbf{79.18} & \textbf{78.84} & \textbf{77.94} & \textbf{78.87} \\
 
    \bottomrule
  \end{tabular}}
\end{table}

\Cref{tab:main} compares \method with intermediate fusion, late fusion and per-frame registration.
Under online poses, intermediate fusion gains little: the three models range from 61.11 to 62.96 \apseven against 62.15 for the ego detector alone, although they do detect more objects at the loose threshold (up to 77.81 \apfive against 70.93).
They find the hidden objects but cannot place them precisely enough.
Late fusion with online poses behaves similarly at \apseven (63.18).
Per-frame registration helps, confirming that geometry is the bottleneck, but each module re-estimates the pose in every frame from the current correspondences: the best (RoCo) reaches 64.36 \apseven, and smoothing the per-frame estimates over time does not close the gap (\cref{sec:robust}).

\method reaches 67.37/66.88/66.54/65.72 \apseven at 0/100/200/400\,ms, 66.63 on average: 3.45 points above late fusion with the same readout and online poses, 3.67 above the best intermediate fusion and 2.27 above the best per-frame registration module.
It also has the highest \apfive of the deployable methods in \cref{tab:main} (78.87 on average), so the gain is not bought by trading recall for precision.
Relative to placing the boxes with the dataset's annotation-assisted offset (69.52, \cref{sec:geometry}), which is not available online, \method closes 54\% of the gap left by online poses without using any annotation.

\subsection{Across detectors}
\label{sec:detectors}

\begin{table}[t]
  \centering
  \caption{\textbf{\method on five detectors.} \apseven (\%) of \method averaged over budgets and readout heads, and its paired difference to late fusion with the online pose (column: online) and to estimating the offset from the current message only (column: single), with the same readout. Every per-head 95\% interval of the difference to the online pose lies above zero; see \cref{tab:population} for intervals.}
  \label{tab:detectors}
  \small
  \setlength{\tabcolsep}{5pt}
  \begin{tabular}{l rrr rrr}
    \toprule
    & \multicolumn{3}{c}{validation} & \multicolumn{3}{c}{held-out} \\
    \cmidrule(lr){2-4}\cmidrule(lr){5-7}
    Detector & \method & $-$online & $-$single & \method & $-$online & $-$single \\
    \midrule
    TraF-Align (released) & 66.63 & +3.45 & +0.57 & 72.57 & +8.37 & +1.38 \\
TraF-Align (strict retrain) & 58.96 & +2.74 & +0.30 & 61.57 & +6.82 & +1.16 \\
V2X-ViT, seed A & 66.08 & +2.38 & +0.49 & 71.30 & +5.92 & +1.24 \\
V2X-ViT, seed B & 64.97 & +2.22 & +0.42 & 68.95 & +4.12 & +0.83 \\
Where2comm & 65.14 & +3.05 & +0.46 & 71.29 & +6.73 & +1.19 \\
 
    \bottomrule
  \end{tabular}
\end{table}

\Cref{tab:detectors} applies \method to the boxes of all five detectors, with $q$ and $\tau_p$ selected per detector on the calibration split and every other setting shared.
On the validation split \method improves on late fusion with online poses by 2.2--3.5 \apseven and on per-message estimation by 0.3--0.6 for every detector; on the held-out split the improvements are larger (4.1--8.4 and 0.8--1.4).
The improvement is not specific to one detector or architecture, and with V2X-ViT or Where2comm boxes \method reaches 65.0--66.1 \apseven, so the result in \cref{tab:main} does not stem from a stronger single-agent detector.

\subsection{Why tracking: temporal aggregation}
\label{sec:temporal}

\begin{table}[t]
  \centering
  \caption{\textbf{How the per-message anchor evidence is aggregated over time.} Released TraF-Align boxes, same anchors and readout, mean over budgets and heads.}
  \label{tab:temporal}
  \small
  \begin{tabular}{l rr rr}
    \toprule
    & \multicolumn{2}{c}{validation} & \multicolumn{2}{c}{held-out} \\
    \cmidrule(lr){2-3}\cmidrule(lr){4-5}
    Estimator of $\boldsymbol\delta_t$ & \apseven & \apfive & \apseven & \apfive \\
    \midrule
    Current message, mean & 66.06 & 78.17 & 71.19 & 85.93 \\
Current message, median & 66.05 & 78.14 & 71.29 & 85.90 \\
Hold last observation & 65.99 & 78.26 & 71.05 & 85.80 \\
Exponential moving average & 66.17 & 78.50 & 72.49 & 87.03 \\
Sliding-window mean & 66.63 & 78.71 & 72.07 & 86.77 \\
\textbf{Common-mode KF (CoMET)} & 66.63 & 78.87 & 72.57 & 86.99 \\
 
    \bottomrule
  \end{tabular}
\end{table}

\Cref{tab:temporal} keeps the anchors fixed and changes only how they are aggregated.
Using the current message alone, by its mean or median, gives 66.06 \apseven; holding the last observation through anchor-free messages does not help.
Pooling over time helps on both splits, and the Kalman filter is the best or tied for best everywhere: it matches the sliding window on the validation split (66.63) and is ahead of it on the held-out split (72.57 against 72.07), where the exponential average is close (72.49).
The filter is preferable to fixed windows for a second reason: its gain adapts to the number of anchors and to the time since the last one, which a window or a constant-gain average cannot do.
On the held-out split the difference between the filter and per-message estimation reaches 1.38 \apseven.

\subsection{Anchors, association and learned alternatives}
\label{sec:ablation}

\begin{table}[t]
  \centering
  \caption{\textbf{Association and anchor choices.} \apseven (\%) averaged over the five detectors, three heads and four budgets. This diagnostic keeps each detector's earlier $q$ and $\tau_p$, so absolute values differ from \cref{tab:main}; the calibration column is the selection criterion. Prediction-gated: remote boxes are shifted by the current estimate before matching.}
  \label{tab:ablation}
  \small
  \setlength{\tabcolsep}{4pt}
  \resizebox{\linewidth}{!}{%
  \begin{tabular}{lllrrr}
    \toprule
    & Anchor candidates & Association & calibration & validation & held-out \\
    \midrule
    N0 & confident ego anchors (score $\ge\tau_p$) & Hungarian + mutual NN & 66.39 & 64.15 & 68.94 \\
U2 & confident ego anchors & Sinkhorn & 65.72 & 64.12 & 68.59 \\
U3 & confident ego anchors & Hungarian, prediction-gated & 66.02 & 64.14 & 68.34 \\
U0 & all ego boxes & Sinkhorn, $\sigma{=}1$\,m, centred on estimate & 66.12 & 64.40 & 68.79 \\
U1 & all ego boxes & Sinkhorn, $\sigma{=}1$\,m, uncentred & 66.28 & 63.92 & 68.04 \\
U5 & all ego boxes & Sinkhorn, $\sigma{=}0.5$\,m & 65.87 & 64.16 & 69.09 \\
U6 & all ego boxes & Sinkhorn, $\sigma{=}2$\,m & 65.92 & 63.91 & 67.45 \\
N2 & all ego boxes & learned soft assignment & 66.24 & 64.41 & 68.69 \\
H2 & all ego boxes & Hungarian, prediction-gated (2.5\,m) & 66.24 & 64.41 & 68.40 \\
H3 & all ego boxes & Hungarian, prediction-gated (1.5\,m) & 66.30 & 64.44 & 68.56 \\
H1 & \textbf{all ego boxes} & \textbf{Hungarian + mutual NN} & 66.89 & 64.36 & 69.06 \\
 
    \bottomrule
  \end{tabular}}
\end{table}

\Cref{tab:ablation} varies the anchor set and the association while keeping the filter.
Two findings matter.
\emph{Use all ego detections.}
Restricting anchors to confident ego detections (N0, U2, U3) loses evidence; with all ego boxes and hard mutual-nearest-neighbour matching (H1, the configuration used throughout) the calibration score is the highest of all variants.
Consistently, raising an anchor score threshold only lowers \apseven (\cref{tab:sensitivity}), and the result is flat for gates between 2.0 and 3.0\,m.
\emph{Structure beats capacity.}
Soft assignment with Sinkhorn normalisation~\cite{cuturi2013sinkhorn} at various widths (U0, U1, U5, U6) and a learned soft assignment trained end to end (N2) do not improve on hard matching: on the validation split all variants lie within 0.6 points of each other, on the calibration split hard assignment is the best, and on the held-out split it is within 0.03 points of the best.
Gating the association with the filter prediction (H2, H3) is worse on the calibration and held-out splits: when the current offset is large, the prediction hides exactly the anchors that would reveal it.
We therefore keep the simplest configuration that wins on the calibration split.

\begin{table}[t]
  \centering
  \caption{\textbf{Sensitivity} (report only, never used for selection). Released TraF-Align boxes, validation \apseven. ``orig.'': the original high-score anchor rule.}
  \label{tab:sensitivity}
  \small
  \setlength{\tabcolsep}{4.5pt}
  \begin{tabular}{l rrrrrrr}
    \toprule
    Anchor score threshold & all & 0.1 & 0.2 & 0.3 & 0.4 & 0.5 & orig. \\
\midrule
AP@0.7 & 66.63 & 66.63 & 66.63 & 66.50 & 66.25 & 65.99 & 65.98 \\
\midrule
Association gate (m) & 1.5 & 2.0 & 2.5 & 3.0 & & & \\
\midrule
AP@0.7 & 66.21 & 66.65 & 66.63 & 66.66 & & & \\
 
    \bottomrule
  \end{tabular}
\end{table}

\subsection{Per-frame registration with temporal smoothing}
\label{sec:robust}

\begin{table}[t]
  \centering
  \caption{\textbf{External registration modules with temporal processing.} Validation \apseven (\%), released TraF-Align boxes, same readout as \method. Median: temporal median over the last messages; gated KF: Kalman filter on the module's per-frame estimate with innovation gating. Window length and gate are selected on the calibration split. \method: 66.63.}
  \label{tab:robust}
  \small
  \begin{tabular}{l rrr}
    \toprule
    Module & per frame & + temporal median & + gated KF \\
    \midrule
    FreeAlign & 63.49 & 63.54 & 63.88 \\
RoCo & 64.36 & 64.30 & 64.20 \\
CBM & 64.06 & 64.00 & 63.88 \\
 
    \bottomrule
  \end{tabular}
\end{table}

One could argue that the advantage of \method comes from temporal smoothing alone, which could be added to any registration module.
\Cref{tab:robust} tests this by smoothing the per-frame estimates of FreeAlign, RoCo and CBM with a temporal median or an innovation-gated Kalman filter, with parameters selected on the calibration split.
Smoothing changes their \apseven by at most 0.4 points; the variant selected on the calibration split (CBM with a three-message median) reaches 64.00, 2.62 points below \method.
A plausible reason is that per-frame registration from object graphs or local context fails grossly when few objects are shared, and smoothing a sequence that contains gross failures does not produce a good estimate, whereas the mutual-nearest-neighbour anchor observation of \method rarely fails grossly (\cref{sec:structure}).

\subsection{\method as the pose of a pose-graph system}
\label{sec:coalign}

\begin{table}[t]
  \centering
  \caption{\textbf{CoAlign with different relative poses.} Full CoAlign fusion, three training seeds, common evaluation region. \apseven per budget and mean, and mean \apfive.}
  \label{tab:coalign}
  \small
  \begin{tabular}{l rrrr r r}
    \toprule
    Relative pose & 0 & 100 & 200 & 400 & mean & \apfive \\
    \midrule
    CoAlign pose graph~\cite{lu2023coalign} & 56.99 & 56.64 & 56.47 & 56.26 & 56.59 & 72.30 \\
Anchor-gated KF & 57.95 & 57.74 & 57.52 & 57.21 & 57.60 & 73.67 \\
\textbf{CoMET} & 58.06 & 57.82 & 57.59 & 57.33 & 57.70 & 73.74 \\
 
    \bottomrule
  \end{tabular}
\end{table}

CoAlign~\cite{lu2023coalign} optimises an agent--object pose graph in every frame before multi-scale feature fusion.
Feeding \method's causal pose to the CoAlign fusion network instead of the pose-graph output improves \apseven by 1.11 points (57.70 against 56.59, consistently for all three seeds) and \apfive by 1.44 (\cref{tab:coalign}).
The gain over an earlier version of our filter that used only confident ego detections as anchors is small (+0.10), which shows that for intermediate fusion the main benefit comes from tracking itself rather than from the choice of anchors.
CoAlign's per-frame graph cannot use evidence from earlier messages; \method can.

\subsection{Robustness to the reference}
\label{sec:reference}

\begin{table}[t]
  \centering
  \caption{\textbf{Results by label-defined population}, validation \apseven, paired differences in points (mean over three heads). ``All'' is the official result; ``co-lab.'' are messages with a co-labelled object, on which the official ground truth of every object is registered; ``zero'' are placeholder messages, on which the official ground truth of roadside-only objects is the uncorrected projection.}
  \label{tab:population}
  \small
  \setlength{\tabcolsep}{3.5pt}
  \resizebox{\linewidth}{!}{%
  \begin{tabular}{l rrr rrr}
    \toprule
    & \multicolumn{3}{c}{\method $-$ online pose} & \multicolumn{3}{c}{\method $-$ single message} \\
    \cmidrule(lr){2-4}\cmidrule(lr){5-7}
    Detector & all & co-lab. & zero & all & co-lab. & zero \\
    \midrule
    TraF-Align (released) & +3.45 & +4.53 & $-$3.42 & +0.57 & +0.98 & $-$2.02 \\
TraF-Align (strict) & +2.74 & +3.47 & $-$2.01 & +0.30 & +0.53 & $-$1.10 \\
V2X-ViT A & +2.38 & +3.16 & $-$2.61 & +0.49 & +0.62 & $-$0.28 \\
V2X-ViT B & +2.22 & +2.93 & $-$2.32 & +0.42 & +0.55 & $-$0.35 \\
Where2comm & +3.05 & +3.97 & $-$2.73 & +0.46 & +0.57 & $-$0.15 \\
 
    \bottomrule
  \end{tabular}}
\end{table}

\paragraph{Populations.}
\Cref{tab:population} splits the official result.
On messages with co-labelled objects, where the official ground truth of every object is registered, \method improves more than on all messages, by 2.9--4.5 points over online poses and 0.5--1.0 over per-message estimation, and all per-head intervals of both differences lie above zero for every detector.
On placeholder messages the official ground truth of roadside-only objects is the uncorrected projection, so moving these boxes to where the objects are is counted as an error there; the auxiliary references below show that this effect is a label artefact.
We therefore do not use any rule that withholds the correction on frames without anchors, which would gain on the official table only through the placeholder.

\begin{table}[t]
  \centering
  \caption{\textbf{Auxiliary references} (sensitivity; not physical truth). \apseven differences with 95\% intervals, mean over three heads. Interpolated: placeholder messages re-registered with the per-sequence interpolation of non-zero offsets (released TraF-Align boxes). Label-median: messages with at least three co-labelled objects re-registered with their median displacement.}
  \label{tab:auxgt}
  \small
  \setlength{\tabcolsep}{4pt}
  \resizebox{\linewidth}{!}{%
  \begin{tabular}{ll lll}
    \toprule
    & & interpolated & \multicolumn{2}{c}{label-median} \\
    \cmidrule(lr){3-3}\cmidrule(lr){4-5}
    Split & Difference & TraF-Align (released) & TraF-Align (released) & TraF-Align (strict) \\
    \midrule
    validation & CoMET $-$ online pose & +4.00 [2.41, 5.69] & +3.93 [2.26, 5.78] & +2.84 [1.66, 4.08] \\
validation & CoMET $-$ current message & +0.94 [0.58, 1.33] & +0.90 [0.50, 1.35] & +0.35 [0.09, 0.60] \\
held-out & CoMET $-$ online pose & +8.73 [2.51, 12.52] & +9.30 [2.65, 13.43] & +7.32 [1.95, 10.57] \\
held-out & CoMET $-$ current message & +1.58 [0.18, 2.58] & +1.42 [0.05, 2.42] & +1.10 [0.16, 1.78] \\
 
    \bottomrule
  \end{tabular}}
\end{table}

\paragraph{Auxiliary references.}
\Cref{tab:auxgt} repeats the main comparisons under the two auxiliary references.
With either reference the improvement over online poses grows (validation: $+4.0$ and $+3.9$ for the released TraF-Align boxes) and the improvement over per-message estimation stays positive with intervals above zero on both splits and both detectors ($+0.35$ to $+1.58$).
On placeholder messages the sign reverses: once the roadside-only objects there are registered, \method is better than the online pose (released $+0.83$, strict $+0.38$ on validation, both intervals above zero; \cref{tab:auxfull}), confirming that the official penalty on these messages is a label artefact.
The conclusions of \cref{tab:main,tab:temporal} therefore do not depend on the label artefacts.

\subsection{Second tier: retraining intermediate fusion}
\label{sec:tier2res}

\begin{table}[t]
  \centering
  \caption{\textbf{TraF-Align retrained from the released weights} (fusion and head fine-tuned for 2,000 updates, reader and backbone frozen, three seeds), validation, mean over budgets and seeds. Top: absolute AP. Bottom: tier 2 minus each alternative, with 95\% intervals.}
  \label{tab:tier2}
  \small
  \setlength{\tabcolsep}{5pt}
  \begin{tabular}{l rr}
    \toprule
    Training and test pose / training labels & \apseven & \apfive \\
    \midrule
    Online pose, original labels & 65.68 & 80.26 \\
Anchor-gated pose, original labels & 66.18 & 81.19 \\
Anchor-gated pose, cleaned labels & 66.44 & 81.36 \\
\textbf{CoMET pose, cleaned labels (tier 2)} & 66.49 & 81.44 \\
\midrule
CoMET without retraining (tier 1) & 66.63 & 78.87 \\
 
    \bottomrule
  \end{tabular}

  \vspace{2mm}
  \begin{tabular}{l ll}
    \toprule
    Tier 2 minus & \apseven & \apfive \\
    \midrule
    retraining with the online pose & +0.81 [0.18, 1.62] & +1.17 [0.58, 1.87] \\
CoMET without retraining (tier 1) & $-$0.14 [$-$1.29, 0.88] & +2.56 [1.12, 3.98] \\
 
    \bottomrule
  \end{tabular}
\end{table}

\Cref{tab:tier2} retrains the fusion network and detection head of TraF-Align for a fixed budget, changing only the pose and, optionally, the training labels.
Compared with retraining under the online pose, tier 2 improves validation \apseven by 0.81 and \apfive by 1.17 points, both with intervals above zero.
Its \apfive of 81.44 is the highest of all methods on the validation split, 2.56 points above \method{} without retraining.
At \apseven the retrained network is on par with \method{} without retraining ($-0.14$, interval including zero), so retraining mainly adds recall.
Retraining under a corrected pose accounts for most of the improvement over the online pose, and cleaned training labels add a further 0.25 \apseven (66.18 to 66.44 with the earlier anchor-gated pose); held-out results are in \cref{app:tier2}.

\begin{table}[t]
  \centering
  \caption{\textbf{The same recipe on two more architectures}, validation, three seeds. \apseven / \apfive of the fusion network retrained with the online pose, with an earlier anchor-gated pose, and with \method poses and cleaned labels (tier 2); last two columns: tier 2 minus the online pose with 95\% intervals.}
  \label{tab:tier2cross}
  \small
  \setlength{\tabcolsep}{3.5pt}
  \resizebox{\linewidth}{!}{%
  \begin{tabular}{l lll ll}
    \toprule
    Architecture & online & anchor-gated & tier 2 & $\Delta$\apseven & $\Delta$\apfive \\
    \midrule
    Where2comm & 63.05 / 76.96 & 63.03 / 77.07 & 63.29 / 77.28 & +0.24 [$-$0.16, 0.54] & +0.32 [$-$0.29, 0.84] \\
CoAlign & 56.47 / 72.13 & 57.85 / 73.78 & 58.10 / 73.88 & +1.63 [1.20, 2.10] & +1.75 [1.03, 2.62] \\
 
    \bottomrule
  \end{tabular}}
\end{table}

\Cref{tab:tier2cross} applies the same recipe to Where2comm and CoAlign.
Tier 2 improves on retraining with online poses for both: clearly for CoAlign ($+1.63$ \apseven, $+1.75$ \apfive, intervals above zero) and slightly for Where2comm ($+0.24$, $+0.32$).
The second tier is thus a drop-in improvement for systems that retrain their fusion network anyway, while the first tier, which needs no retraining, remains the default.

\subsection{Controlled study on OPV2V}
\label{sec:opv}

\begin{table}[t]
  \centering
  \caption{\textbf{OPV2V with controlled partner pose error.} \apseven (\%), mean over three readout heads. Stale: remote boxes placed with the perturbed pose; oracle: the true pose.}
  \label{tab:opv}
  \small
  \setlength{\tabcolsep}{4pt}
  \begin{tabular}{l rrr rrr}
    \toprule
    & \multicolumn{3}{c}{$d=0$} & \multicolumn{3}{c}{$d=400$\,ms} \\
    \cmidrule(lr){2-4}\cmidrule(lr){5-7}
    Pose error & stale & \method & oracle & stale & \method & oracle \\
    \midrule
    structured, fitted to V2X-Seq & 73.22 & 82.14 & 85.07 & 66.90 & 70.73 & 74.62 \\
i.i.d.\ per frame (0.2\,m, 0.2$^\circ$) & 80.50 & 81.10 & 85.07 & 67.43 & 69.19 & 74.62 \\
ego pose time error 50\,ms & 77.70 & 80.41 & 85.07 & 67.91 & 72.58 & 74.62 \\
ego pose time error 100\,ms & 70.82 & 76.36 & 85.07 & 67.64 & 71.35 & 74.62 \\
ego pose time error 150\,ms & 68.58 & 74.42 & 85.07 & 67.56 & 70.49 & 74.62 \\
 
    \bottomrule
  \end{tabular}
\end{table}

In simulation we can choose the error.
\Cref{tab:opv} perturbs the partner pose of OPV2V~\cite{xu2022opv2v} in three ways.
With a structured error whose statistics are fitted to V2X-Seq, \method recovers 8.9 of the 11.9 \apseven points lost at $d=0$ and 3.8 of 7.7 at 400\,ms.
With independent per-frame noise, which has no persistent component, the stale boxes are already close to the oracle and \method still adds 0.6 and 1.8 points.
With an ego-pose timing error, which produces a persistent offset during motion, \method recovers 2.7--5.8 points at $d=0$.
\method helps in proportion to how persistent and how shared the error is.

\section{Discussion}
\label{sec:discussion}

\paragraph{Reading V2X-Seq results with online pose correction.}
Our audit changes how such results should be read.
The dataset offset is the right reference for the official ground truth, because the ground truth is built from it; it is not a measurement of the physical pose error, being zero where it could not be estimated and clipped at the bounds of its search.
A method that estimates the true error is therefore counted as wrong on placeholder frames and on saturated frames.
Population splits and auxiliary references do not replace physical ground truth, but they show which conclusions depend on these artefacts; for \method, none of these conclusions changes.
We hope that future releases document how such offsets are produced and flag placeholders explicitly.

\paragraph{Limitations.}
\method assumes that one translation dominates the error of a message and that the objects seen by both agents share it with those seen only by the partner.
When this does not hold, for example when the partner's objects are affected by errors that the co-visible objects do not reflect, the transferred correction can misplace boxes; a consistency check on the normalised filter innovation is a natural safeguard.
The model is planar and translation-only, and the estimate follows abrupt changes of the offset with a short delay.
Because anchors are gated at 2.5\,m, an offset larger than the gate cannot be acquired from a cold start, although an error that grows after tracking has begun is still followed.
Anchor weighting by object motion or distance and an explicit rotation term are natural extensions.
Our real-data evaluation uses one V2I dataset, complemented by a controlled simulation, and all numbers come from offline replay of recorded messages.

\section{Conclusion}
\label{sec:conclusion}
On real V2I data with online poses and strict arrival, the main obstacle to cooperative detection is a slowly drifting translation shared by all objects of a message, while the age of the message costs far less.
Tracking this translation from co-visible anchors with a causal scalar filter, and transferring it to the objects only the partner sees, turns box messages into a clear gain over intermediate fusion for every detector we tested, without training the estimator; the same pose also improves intermediate fusion when it is retrained.
We further find that the benchmark's own reference offset is a placeholder on a quarter of the frames and saturated on almost half of the rest, which matters for evaluating any geometric correction.
The estimator, protocol and experimental results accompany the manuscript at
\url{https://github.com/sswun/CoMET-V2X}. Use of the code or results requires
prior written permission from the authors. The repository gives the terms and contact details.

\bibliographystyle{plainnat}
\bibliography{references}

\clearpage
\appendix
\section*{Appendix}
\addcontentsline{toc}{section}{Appendix}

\section{Proof of the scalar recursion}
\label{app:proof}
Let the state $\boldsymbol\delta_t\in\mathbb R^2$ follow $\boldsymbol\delta_t=\boldsymbol\delta_{t-1}+\mathbf w_t$ with $\mathbf w_t\sim\mathcal N(\mathbf 0,q^2\mathbf I)$, and let a message with $n_t\ge1$ anchors provide $\bar{\mathbf z}_t=\boldsymbol\delta_t+\bar{\boldsymbol\eta}_t$ with $\bar{\boldsymbol\eta}_t\sim\mathcal N(\mathbf 0,\tfrac{r^2}{n_t}\mathbf I)$; the mean of $n_t$ independent residuals with covariance $r^2\mathbf I$ has covariance $\tfrac{r^2}{n_t}\mathbf I$, and it is a sufficient statistic for $\boldsymbol\delta_t$ because the likelihood of the individual residuals depends on them only through their mean.
Assume $\mathbf P_{t-1}=p_{t-1}\mathbf I$, which holds for $t=1$ with $p_0=1$\,m$^2$.
The prediction gives $\mathbf P_t^-=(p_{t-1}+q^2)\mathbf I=p_t^-\mathbf I$.
With observation matrix $\mathbf I$ the Kalman gain is $\mathbf K_t=\mathbf P_t^-(\mathbf P_t^-+\tfrac{r^2}{n_t}\mathbf I)^{-1}=\tfrac{p_t^-}{p_t^-+r^2/n_t}\mathbf I=k_t\mathbf I$, and the update gives $\mathbf P_t=(\mathbf I-\mathbf K_t)\mathbf P_t^-=(1-k_t)p_t^-\mathbf I$.
Without anchors there is no update and $\mathbf P_t=\mathbf P_t^-$.
By induction the covariance stays isotropic and \cref{eq:gain,eq:update} are exact.
The steady state of \cref{eq:steady} follows from the fixed point $p^-=(1-k)p^-+q^2$ with $k=p^-/(p^-+r^2/n)$, which gives $(p^-)^2-q^2p^--q^2r^2/n=0$.

\section{Audit of the dataset offset}
\label{app:audit}

\paragraph{Procedure.}
For every training and validation frame pair we read the raw vehicle-side and roadside annotations, project the roadside labels into the vehicle frame with the public calibration, and join them to the cooperative labels by track and token.
Cooperative labels carry a source flag (vehicle-only, roadside-only, co-labelled).
For roadside-only objects we compare the cooperative label with the public projection, separately for frames with zero and non-zero offset; a second implementation repeats the join and the inverse transform chain independently.
For co-labelled objects we compute the displacement between the vehicle label and the projected roadside label; its median over the objects of a frame (frames with at least three co-labelled objects) is the label-median displacement used in \cref{sec:audit}.

\paragraph{Findings in detail.}
The roadside-only cooperative label equals the public projection plus the offset, with a residual of about $4\times10^{-7}$\,m, for zero and non-zero offsets alike.
\Cref{tab:census} gives the census; in the fit, held-out and validation splits no zero message contains a co-labelled object, and only 10--16 non-zero messages lack one.
The non-zero values per axis are
$\{-1.10,-1.06,-1.02,-0.98,-0.94\}$, $\{-0.70,\dots,-0.54\}$, $\{-0.30,\dots,-0.14\}$, $\{+0.10,\dots,+0.26\}$ and $\{+0.50,\dots,+0.66\}$ in steps of 0.04\,m, five values per group; together with zero these are 26 values.
For interior values the median label-median displacement of the same frames differs from the value by at most about 0.1\,m (\cref{fig:audit}); on the bounds the displacement continues beyond the bound.
Over frames with at least three co-labelled objects and a non-zero offset, the correlation between offset and label-median displacement is 0.85 for $x$ and 0.92 for $y$, and their median difference is 0.16\,m; the offset is therefore close to, but not equal to, the label displacement, and we do not interpret it as a mean.

\section{Implementation details}
\label{app:impl}

\paragraph{Hyper-parameters.}
$r=0.6$\,m, $g=2.5$\,m, $p_0=1$\,m$^2$ for all detectors.
Selected on the calibration split: $q=0.05$/$0.025$/$0.1$/$0.05$/$0.05$\,m and $\tau_p=0.5$/$0.3$/$0.5$/$0.5$/$0.5$ for the released TraF-Align, the strictly retrained TraF-Align, the two V2X-ViT models and Where2comm.
The candidate grid for $q$ was $\{0.025,0.05,0.1,0.15\}$\,m.
The choice between the original readout heads and heads refitted on \method geometry was made once for all detectors on the calibration split (mean over five detectors 65.70 vs.\ 65.66 \apseven for heads refitted after removing placeholder frames; original heads kept).

\paragraph{Anchor statistics.}
\Cref{tab:anchors} lists the mean number of anchors per message and the fraction of messages without anchors for the released TraF-Align boxes.

\begin{table}[h]
  \centering
  \caption{Mean anchors per message / fraction of messages without anchors, released TraF-Align boxes.}
  \label{tab:anchors}
  \small
  \begin{tabular}{l rrrr}
    \toprule
    Split & 0\,ms & 100\,ms & 200\,ms & 400\,ms \\
    \midrule
    fit & 3.31 / 29.8\% & 3.28 / 30.0\% & 3.20 / 31.1\% & 2.98 / 33.1\% \\
calibration & 3.03 / 33.5\% & 2.97 / 34.6\% & 3.00 / 35.0\% & 2.89 / 34.2\% \\
held-out & 3.34 / 31.7\% & 3.35 / 30.8\% & 3.30 / 31.2\% & 3.11 / 32.3\% \\
validation & 2.96 / 38.2\% & 2.92 / 38.7\% & 2.90 / 37.6\% & 2.73 / 38.5\% \\
 
    \bottomrule
  \end{tabular}
\end{table}

\paragraph{Readout.}
The 39 receiver-side features of a candidate are: message age, detector score, box size, heading in the receiver and source frames, distance and BEV overlap to the nearest ego detection and that detection's score, the number of ego boxes within 10\,m, flags for ego, propagated and causally associated candidates, propagation horizon, time between the two messages used for the velocity, speed, association distance (absolute and relative to the gate) and propagation shift. The 13 features that encode absolute position (receiver and source coordinates and ranges, and the source-to-receiver translation and range) are zeroed.
The head is a two-layer perceptron (39--32--1, 1,313 parameters) trained with binary cross-entropy on candidates of the fit split; three heads with seeds 2023/2024/2025 are trained and all results average over them.

\paragraph{Second tier.}
Initial weights: the released TraF-Align weights (epoch 60).
Reader and backbone are frozen (batch normalisation in evaluation mode); the fusion network, multi-scale aggregation and detection head are trained.
AdamW, learning rate $10^{-4}$, weight decay 0.01, gradient accumulation 4, 2,000 updates (8,000 micro-batches, about one pass over the 7,524 fit conditions at $d\in\{0,400\}$\,ms), no checkpoint selection on validation data.
The loss is the original anchor classification and box regression loss on current cooperative ground truth.
The same budget is used for the online-pose and \method arms and for Where2comm and CoAlign.

\section{Auxiliary references in detail}
\label{app:aux}

\begin{table}[h]
  \centering
  \caption{\textbf{Interpolated reference with three gap limits} (released TraF-Align boxes, \apseven, 95\% intervals). Placeholder runs are re-registered only if they are at most 1\,s or 3\,s long, or always.}
  \label{tab:auxinterp}
  \small
  \setlength{\tabcolsep}{4pt}
  \resizebox{\linewidth}{!}{%
  \begin{tabular}{l ll ll}
    \toprule
    & \multicolumn{2}{c}{validation} & \multicolumn{2}{c}{held-out} \\
    \cmidrule(lr){2-3}\cmidrule(lr){4-5}
    Re-registered & \method $-$ online & \method $-$ single & \method $-$ online & \method $-$ single \\
    \midrule
    gaps $\le$1\,s & +3.66 [1.93, 5.48] & +0.60 [0.06, 1.13] & +8.43 [1.80, 12.41] & +1.43 [$-$0.15, 2.53] \\
gaps $\le$3\,s & +3.66 [1.93, 5.48] & +0.60 [0.06, 1.13] & +8.45 [1.89, 12.45] & +1.45 [$-$0.11, 2.55] \\
all gaps & +4.00 [2.41, 5.69] & +0.94 [0.58, 1.33] & +8.73 [2.51, 12.52] & +1.58 [0.18, 2.58] \\
 
    \bottomrule
  \end{tabular}}
\end{table}

\begin{table}[h]
  \centering
  \caption{\textbf{Label-median reference by population} (\apseven differences, mean over three heads; every 95\% interval excludes zero). On placeholder messages (zero) \method is now better than the online pose as well, because the reference registers the roadside-only objects.}
  \label{tab:auxfull}
  \small
  \setlength{\tabcolsep}{3pt}
  \resizebox{\linewidth}{!}{%
  \begin{tabular}{ll rrr rrr}
    \toprule
    & & \multicolumn{3}{c}{\method $-$ online} & \multicolumn{3}{c}{\method $-$ single} \\
    \cmidrule(lr){3-5}\cmidrule(lr){6-8}
    Detector & Split & all & co-lab. & zero & all & co-lab. & zero \\
    \midrule
    TraF-Align (released) & val. & +3.93 & +4.45 & +0.83 & +0.90 & +0.94 & +0.77 \\
TraF-Align (released) & held-out & +9.30 & +10.21 & +0.58 & +1.42 & +1.56 & +0.22 \\
TraF-Align (strict) & val. & +2.84 & +3.24 & +0.38 & +0.35 & +0.40 & +0.13 \\
TraF-Align (strict) & held-out & +7.32 & +8.10 & +0.58 & +1.10 & +1.22 & +0.22 \\
 
    \bottomrule
  \end{tabular}}
\end{table}

\section{Second tier in detail}
\label{app:tier2}

\paragraph{Held-out split.}
\Cref{tab:tier2heldout} gives the absolute AP of the retrained TraF-Align models on the held-out split (inside the detectors' training data). Tier 2 improves on retraining with online poses by $+3.15$ \apseven $[0.23, 4.81]$ and $+1.91$ \apfive.

\begin{table}[h]
  \centering
  \caption{\textbf{Retrained TraF-Align on the held-out split}, mean over budgets and three seeds.}
  \label{tab:tier2heldout}
  \small
  \begin{tabular}{l rr}
    \toprule
    Training and test pose / training labels & \apseven & \apfive \\
    \midrule
    Online pose, original labels & 68.30 & 87.68 \\
Anchor-gated pose, original labels & 72.15 & 90.09 \\
Anchor-gated pose, cleaned labels & 72.73 & 89.94 \\
\textbf{CoMET pose, cleaned labels (tier 2)} & 71.45 & 89.59 \\
\midrule
CoMET without retraining (tier 1) & 72.57 & 86.99 \\
 
    \bottomrule
  \end{tabular}
\end{table}

\paragraph{Label variants.}
Variant A re-registers roadside-only training labels on placeholder messages with the per-sequence interpolation of non-zero offsets.
Variant B additionally moves roadside-only labels on axes where the offset sits on a search bound to the label-median displacement of the frame (frames with at least three co-labelled objects).
On the calibration split A scored 67.91 and B 66.66 \apseven (three seeds, four budgets), so A was selected before the validation and held-out splits were evaluated; \cref{tab:tierab} reports both.
The calibration split has the highest share of saturated messages (77.7\%), on which the official ground truth favours predictions placed at the bound, which explains why the selection prefers A; we keep the variant selected in advance.

\begin{table}[h]
  \centering
  \caption{\textbf{Label variants A (selected) and B} minus retraining with online poses, TraF-Align, three seeds, 95\% intervals.}
  \label{tab:tierab}
  \small
  \setlength{\tabcolsep}{3pt}
  \resizebox{\linewidth}{!}{%
  \begin{tabular}{l ll ll}
    \toprule
    & val.\ \apseven & val.\ \apfive & held-out \apseven & held-out \apfive \\
    \midrule
    A $-$ online-pose fine-tune & +0.81 [0.18, 1.62] & +1.17 [0.58, 1.87] & +3.15 [0.23, 4.81] & +1.91 [$-$0.67, 3.05] \\
B $-$ online-pose fine-tune & +1.08 [0.37, 1.94] & +1.40 [0.81, 2.10] & +3.74 [0.55, 5.38] & +2.13 [$-$0.02, 3.12] \\
 
    \bottomrule
  \end{tabular}}
\end{table}

\paragraph{Selection criterion fixed in advance.}
Before evaluating the validation split we required, for the selected variant, that the 95\% intervals of the difference to retraining with the online pose lie above zero for both \apseven and \apfive, and that the difference to \method{} without retraining be at least $+1.0$ \apfive with an interval above zero and no worse than $-0.30$ \apseven.
All five conditions were met.

\begin{table}[h]
  \centering
  \caption{\textbf{Second tier under the auxiliary references}: tier 2 minus retraining with online poses (TraF-Align, 95\% intervals).}
  \label{tab:tier2aux}
  \small
  \setlength{\tabcolsep}{2.5pt}
  \resizebox{\linewidth}{!}{%
  \begin{tabular}{ll ll ll}
    \toprule
    Reference & Difference & val.\ \apseven & val.\ \apfive & held-out \apseven & held-out \apfive \\
    \midrule
    interpolated & tier 2 $-$ online-pose fine-tune & +1.19 [0.54, 2.00] & +1.79 [1.11, 2.57] & +3.43 [0.90, 4.92] & +2.40 [0.46, 3.26] \\
label-median & tier 2 $-$ online-pose fine-tune & +1.27 [0.56, 2.07] & +2.29 [1.25, 3.72] & +4.00 [1.26, 5.66] & +3.25 [1.52, 4.15] \\
 
    \bottomrule
  \end{tabular}}
\end{table}

\section{Reproducibility}
\label{app:repro}
The repository contains (i) the stand-alone estimator with unit tests and a script that compares it with the reference per-message states; (ii) a snapshot of the research code that produced every reported result, with a SHA-256 manifest; (iii) every table of this paper in machine-readable form with the source of each number; and (iv) the scripts that generate the tables and figures of this manuscript from those files.
No dataset content, per-frame annotation-derived data or model weights are redistributed.

\end{document}